\documentclass[lettersize,journal]{IEEEtran}
\usepackage{amsmath,amsfonts}
\usepackage{algorithmic}
\usepackage{array}
\usepackage[caption=false,font=normalsize,labelfont=sf,textfont=sf]{subfig}
\usepackage{textcomp}
\usepackage{stfloats}
\usepackage{url}
\usepackage{verbatim}
\usepackage{graphicx}
\usepackage{xspace}
\usepackage{subfloat}  
\usepackage{makecell}
\usepackage{booktabs}
\usepackage{float}
\usepackage{multirow}
\usepackage{array}
\pdfoutput=1  

\def\BibTeX{{\rm B\kern-.05em{\sc i\kern-.025em b}\kern-.08em
    T\kern-.1667em\lower.7ex\hbox{E}\kern-.125emX}}
\usepackage{balance}

\newcommand{\systemname}{{\sf TrajMatch} \xspace}
\newcommand{\simdataset}{{\sf LiDARnet-sim 1.0} \xspace}

\begin{document}
\title{{ TrajMatch: Towards Automatic Spatio-temporal Calibration for Roadside LiDARs
through Trajectory Matching }}

\author{Haojie~Ren, Sha~Zhang, Sugang~Li, Yao~Li, Xinchen~Li,
 Jianmin~Ji, Yu~Zhang,  \\ Yanyong~Zhang,~\IEEEmembership{Fellow,~IEEE}

\thanks{H. Ren,  S. Zhang, Y. Li, X. Li, Y. Zhang, J. Ji and Y. Zhang are with School of Computer Science and Technology, University of Science and Technology of China, Hefei, China (e-mail: rhj@mail.ustc.edu.cn, zhsh1@mail.ustc.edu.cn, zkdly@mail.ustc.edu.cn, yuzhang@ustc.edu.cn, jianmin@ustc.edu.cn, yanyongz@ustc.edu.cn).

S. Li is with Google Cloud, Sunnyvale, Califonia, U.S.A. (e-mail: sugangli@google.com)

Corresponding author: Yanyong Zhang. }
}


\maketitle

\begin{abstract}
Recently, it has become popular to deploy sensors such as LiDARs on the roadside to monitor the passing traffic and assist autonomous vehicle perception. Unlike autonomous vehicle systems, roadside sensors are usually affiliated with different subsystems and lack synchronization both in time and space. Calibration is a key technology which allows the central server to fuse the data generated by different location infrastructures, which can deliver improve the sensing range and detection robustness. Unfortunately, existing calibration algorithms often assume that the LiDARs are significantly overlapped or that the temporal calibration is already achieved. Since these assumptions do not always hold in the real world, the calibration results from the existing algorithms are often unsatisfactory and always need human involvement, which brings high labor costs. 

In this paper, we propose \systemname - the first system that can automatically calibrate for roadside LiDARs in both time and space. The main idea is to automatically calibrate the sensors based  on the result of the detection/tracking task instead of extracting special features. More deeply, we propose a mechanism for evaluating calibration parameters that is consistent with our algorithm, and we demonstrate the effectiveness of this scheme experimentally, which can also be used to guide parameter iterations for multiple calibration. Finally, to evaluate the performance of \systemname, we collect two dataset, one simulated dataset \simdataset and a real-world dataset. Experiment results show that \systemname can achieve a spatial calibration error of less than $10cm$ and a temporal calibration error of less than $1.5ms$.
\end{abstract}

\begin{IEEEkeywords}
roadside traffic monitoring, spatio-temporal calibration, trajectory matching
\end{IEEEkeywords}


\section{INTRODUCTION}

\begin{figure}[!t]
\centering
\includegraphics[width=.75\linewidth]{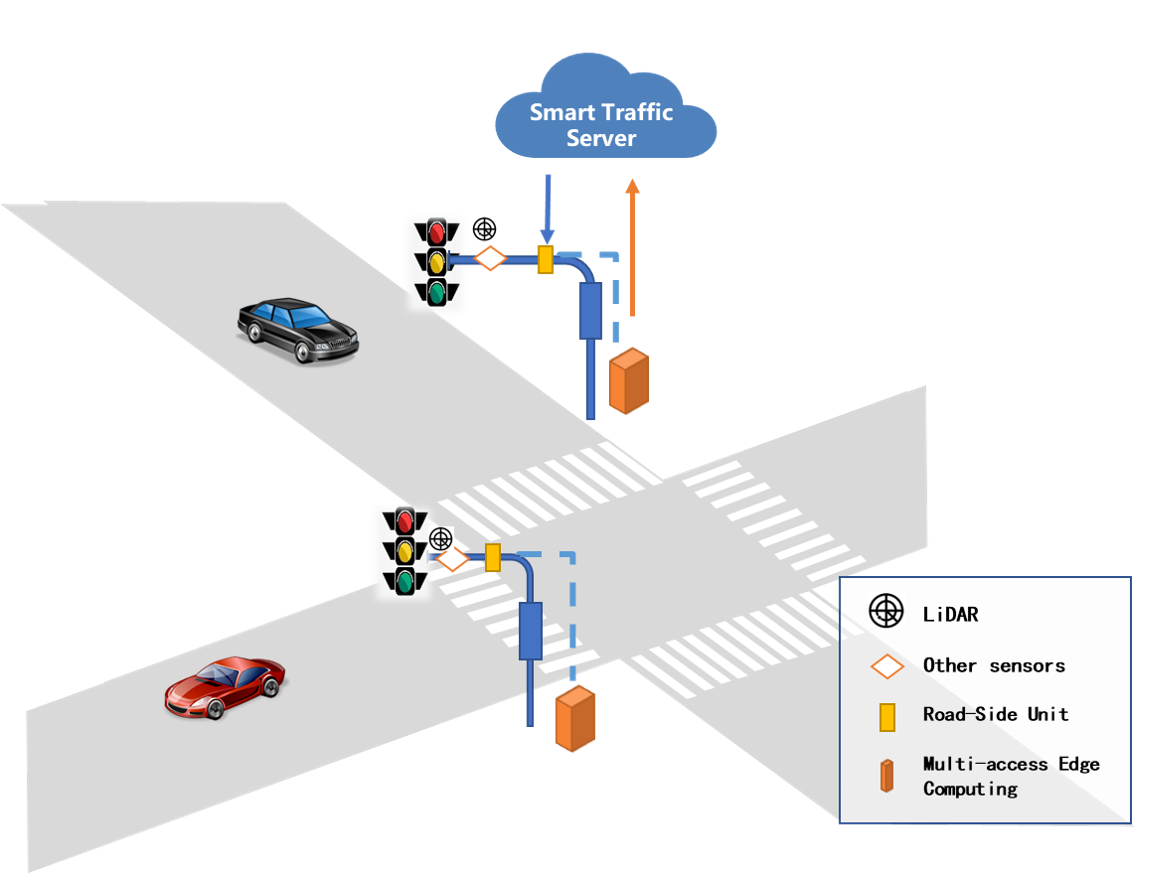}
\caption{ \label{fig:scenario} An example deployment of the roadside traffic monitoring system. We show a classic deployment scenario at an intersection, where LiDARs are installed diagonally for collaborative perception. }
\vspace{-0.2cm}
\end{figure}

\IEEEPARstart{I}{t} has become a recent trend to have smart traffic systems involve cameras/LiDARs at important locations especially at the busy intersections to closely monitor the traffic for improved traffic flow, driving safety and energy efficiency. Specifically, LiDARs are increasingly deployed for such purposes due to its ability to provide accurate ranging information that is particularly important for traffic monitoring. As the accuracy requirement for traffic monitoring continues to go up, it has also become a common practice to deploy multiple sensors at an intersection to collaboratively obtain better coverage~\cite{arnold2020cooperative}. We show an intersection scene in Figure~\ref{fig:scenario}, where intelligent roadside infrastructures equipped with computing and sensing units to capture real-time environment data and perform necessary calculations such as detection and tracking, finally they will transfer the data to a smart traffic server(STS), where their data will be fused. In recent years, LiDAR technologies in detection and tracking has gradually matured and can be easily deployed to roadside infrastructures. In this work, we focus on another key problem of spatio-temporal calibration for the sensors on the infrastructure, since it allows the traffic server to accuracy fusion the data from different location LiDARs for improve the sensing capability for increased sensing range, more robust object detection/tracking and generated a more comprehensive view of the traffic.

We note that there are few studies on calibration of multiple LiDARs. \cite{kim2021automated} Although there exist methods in this topic, but they don't work for roadside multiple LiDARs because of their own limitations and low algorithmic performance. Most studies have focused on the calibration of the two sensors in space, and assumed that the time between sensors have been synchronized ~\cite{kim2021automated, xie2018infrastructure, quenzel2016robust}. For the time dimension, \cite{glas2015snapcat} used Random sample consensus(RANSAC) algorithm to align the trajectories of the LiDARs, but they assumed at most one person can be seen. The same assumption also appears in \cite{6385620}. Obviously, this assumption does not always hold in roadside traffic scenarios. For the space dimension, there are mainly two types of methods, the target-based methods. and the target-less method. The target-based methods require identifiable objects (such as checkerboards, polygonal boards and apriltags), and estimate the relative pose between sensors by aligning the target positions observed on each sensor ~\cite{pusztai2017accurate, xue2019automatic, xie2018infrastructure}. These methods require pricey and lengthy manual  operations, and cannot be applied to large-scale deployment. As for the method of targetless methods, it is usually to extract the features in the environment. This methods cannot achieve high precision in traffic scenes, where exist a large number of repeated structures and lack of distinctive features~\cite{vi-eye}.


With the above challenges in mind, this paper propses \systemname - the first system that can automatically calibrate LiDARs on the roadside both temporally and spatially for supporting various smart traffic applications. The main idea is to use the trajectory information of moving objects provided by the point cloud based object detection/tracking module to search for correspondences between trajectories from different LiDARs. Taking into account the accuracy and robustness of the algorithm, we use a 
three-stage framework from local to global. In first stage, we extract rotation- and translation-invariant motion features at each position where and when an object appears, and establish the position-wise correspondences based upon the feature distances. In second stage, we take into consideration much more discriminative semantics information about the objects as well as their neighborhood distributions, we called middle-level information, to filter the error correspondence. Finally, we iterative optimization process based the whole data to achieve the efficient calibration. 

In order to evaluate the performance of \systemname, we collect a simulation dataset(\simdataset) and a real-world traffic dataset, both for typical 4-way intersections. As for spatial calibration,  the results show that our \systemname can achieve centimeter-level accuracy with a success rate above $90 \%$. As for temporal calibration, our \systemname can achieve less than 1.5 millisecond accuracy.

In summary, the main contributions of this paper are as follows:
\begin{enumerate}
 \item We propose \systemname, the first system that can automatically achieve accurate spatio-temporal calibration for roadside LiDARs. \systemname has the following  key advantages. Firstly, the algorithm  uses the results of the object detection/tracking module to perform LiDAR calibration, without extracting special features for the calibration task separately. Secondly, the algorithm utilizes the spatio-temporal information embedded in object motions, without requiring any features in the environment. Which means that the effect of the algorithm does not depend on the specific environment and has good transferability. Thirdly, it does not assume any initial state about the system, including the location and rotation of each LiDAR and the spatial/temporal distance between them. As such, our system offers a lightweight, automated, efficient and robust LiDAR calibration approach. 

\item We propose a continuous calibration method that can continue to refine the calibration as cars pass by. In particular, our system is able to self-assess the quality of each calibration session and integrate the well-performed sessions to continuously improve the calibration accuracy. 

\item Our system can achieve centimeter-level spatial calibration accuracy as well as millisecond-level temporal calibration accuracy without relying on any external information about the deployment configuration and the environment. 
\end{enumerate}

\section{Background and Motivation} \label{sec:background}

\def\svgwidth{\linewidth}

\begin{figure}[!t]
\centering
\includegraphics[width=.8\linewidth]{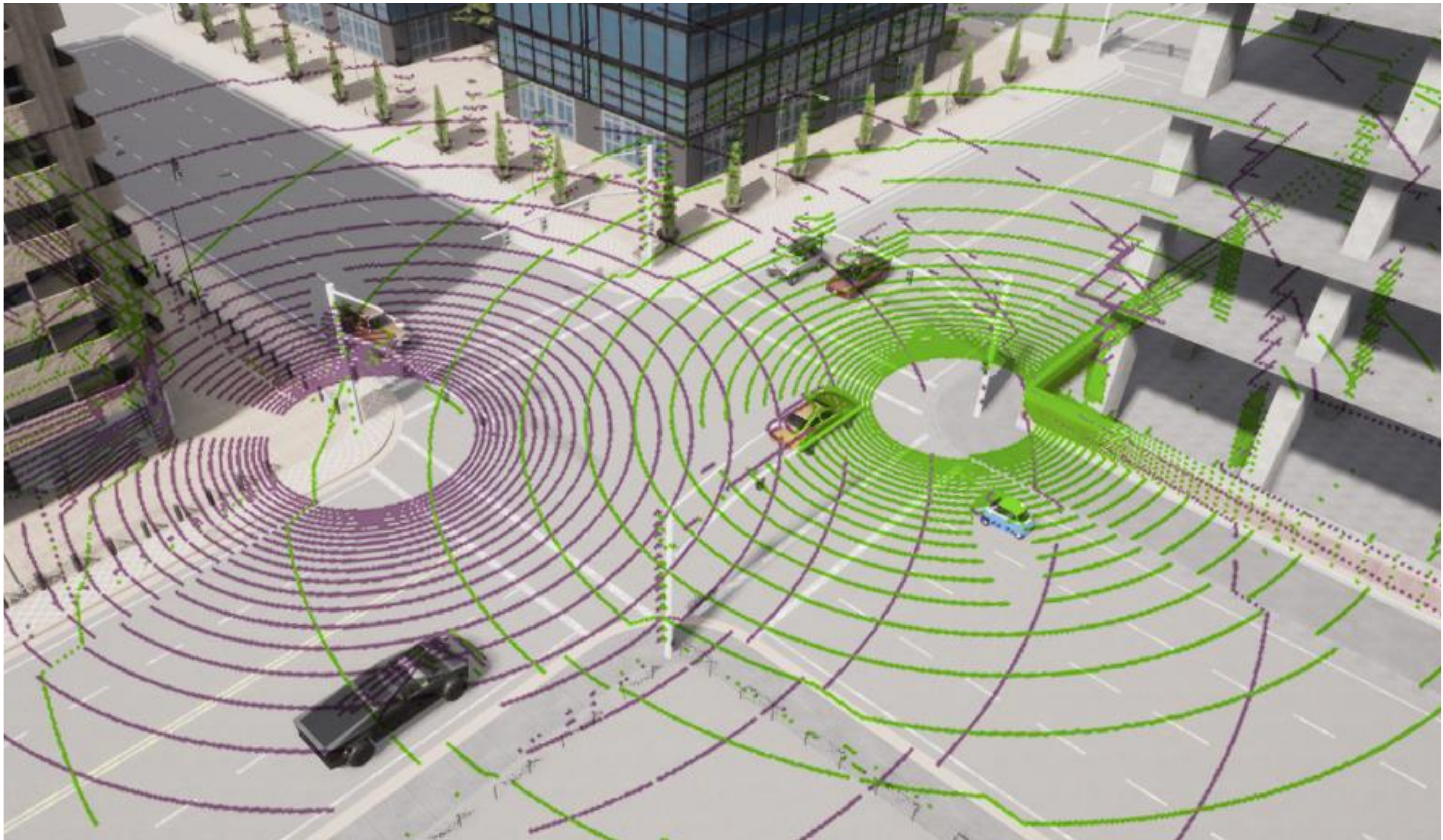}
\caption{\textbf{The point clouds from the two LiDARs diagonally deployed in a 4-way intersection.}}
\label{fig:low_overlap}
\vspace{-0.4cm}
\end{figure}

\subsection{Roadside Traffic Monitoring Scenarios}

In this paper, we focus on the scenario illustrated in Figure~\ref{fig:scenario}. The LiDAR sensors are installed on the roadside to collaboratively monitor the passing traffic. In such a multi-LiDAR collaborative sensing system, their detection areas have a certain degree of overlap. In Figure~\ref{fig:install} we enumerate three deployment scenarios: (a) a 4-way intersection; (b) a 3-way intersection; (c) a sidewalk. These LiDARs are not deployed on the same spot and have a significant transformation due to the distances. Additionally, they are usually affiliated with different subsystems, leading to non-synchronized clocks. Sometimes, their time deviation can be as large as a few seconds\cite{Radar-Camera}\cite{WU2091105}. It is worth mentioning that we evaluated the performance of the algorithm in these scenarios and achieved good results  in Sec.~\ref{sec::result}. 

The workflow of the collaborative traffic sensing system usually consists of the following steps: (1) data acquisition, in which the LiDAR sensors obtain point cloud data, (2) object detection, in which the detection module is performed to obtain the bounding boxes of the objects, (3) multi-object tracking, in which the tracking module is performed to generate the tracks for the objects by assigning  track IDs  to  distinguish the trajectories of different objects, (4) upload data to central server, (5) merge the data from different subsystem, which need to know the relative position between subsystems. In this paper, we devise \systemname to perform the  spatio-temporal calibration by matching the trajectories of the same object from different LiDARs. Naturally, it servers for the center server to fusion the data.
%

\vspace{-0.3cm}
\subsection{Problem Formulation}
Next, we formally define the LiDAR spatio-temporal calibration problem. We suppose $P= \left\{ p_1, p_2, p_3, ... \right\} $ and $Q=\left\{q_1, q_2, q_3, ...\right\}$ are two groups of 4D point clouds with time dimension. Points in $P$ and $Q$ have similar forms:
\begin{equation}
    p_i = \left\{ x_i^p, y_i^p, z_i^p, t_i^p \right\}, \;
    q_j = \left\{ x_j^q, y_j^q, z_j^q, t_j^q \right\},
\end{equation}
where $x$, $y$, and $z$ represent the coordinates of the three dimensions in their respective coordinate systems, and $t$ is the timestamp. Especially, before calibration, $P$ and $Q$ have different coordinate systems. For example, the $x$ axis of $P$ may correspond to the $y$ axis of $Q$. Additionally, there may be a certain deviation in the time dimension $t$ for the point clouds collected at the same time.



In this work, we define the problem in the 4D space, for a point $ p = 
\begin{bmatrix}
 x, y, z, t
\end{bmatrix}
^T$, the coordinate transformation can be defined as:
\begin{equation}
    p' = 
    \begin{bmatrix}
    R_3 & 0 \\ 
    0 & 1
    \end{bmatrix}
    \times
    \begin{bmatrix}
        x \\ 
        y \\
        z \\
        t
    \end{bmatrix}
    +
    \begin{bmatrix}
        t_x \\ 
        t_y \\
        t_z \\
        d_t
    \end{bmatrix}
    = R \times p  + T,
\end{equation}
where $d_t$ represents the offset in time, and $\mathbf{R}$ and $\mathbf{T}$ denote the rotation matrix and the translation vector between LiDARs, respectively.

The spatio-temporal calibration problem for two 4D point clouds $P$ and $Q$ can be described as:
\begin{equation}
    R,T = \mathop{\arg\min}_{R,T} \ {\frac{1}{|P|}} \ { \sum_{i=1}^{|P|} {\left \| p_i - (R \times q_i + T) \right \|}^2 },
 \label{equ::least_sqrt}   
\end{equation}
where $p_i$ and $q_i$ are the corresponding points in the $P$ and $Q$. 

In this way, we transform the LiDAR calibration problem into an optimization problem in which we try to find the rotation parameter $R$ and the translation vector $T$, with the objective of minimizing the matching error between the data of the two LiDARs in both time and space. We call the collection of $R$ and $T$ as the system's transformation parameters.


\section{System Overview}\label{sec:overview}  


\begin{figure}[!t]
\centering
\includegraphics[width=\linewidth]{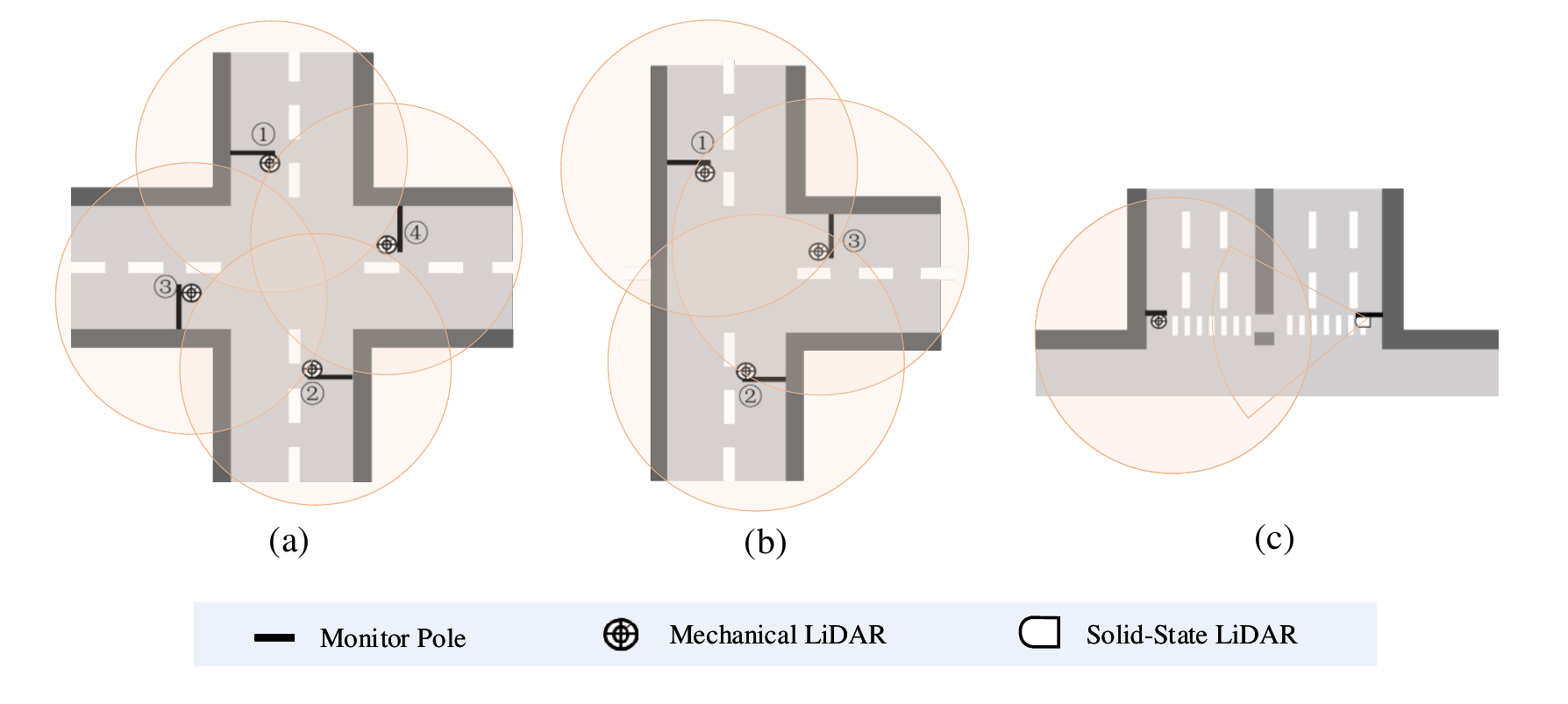}
\caption{\textbf{Typical deployment scenarios for roadside traffic monitoring: (a) a 4-way intersection, (b) a 3-way intersection, and (c) a sidewalk.}}
\label{fig:install}
\end{figure}





\def\svgwidth{\linewidth}
\begin{figure*}[t]
	\centering
 	\includegraphics[width=1.0\linewidth]{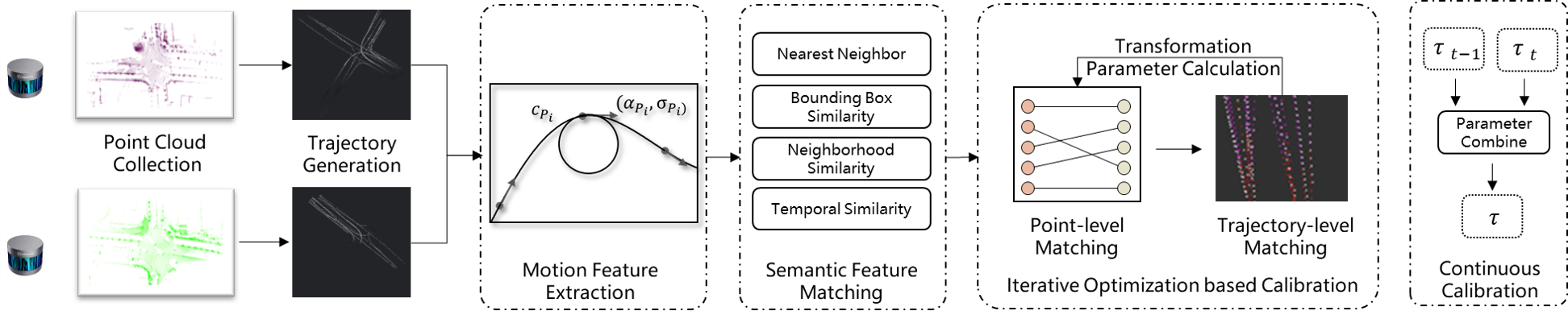}
	\caption{\textbf{Framework of our \systemname system}. In this paper, we exploit a system for performing spatio-temporal calibration for roadside LiDARs. Our system takes LiDARs detection and tracking results as input, and consists of the following main modules: (1) motion feature extraction (\textbf{local}), (2) semantic feature matching (\textbf{middle}), (3) iterative optimization (\textbf{global}), and (4) continuous calibration. Finally, we propose a strategy to adjust the errors over the continuous calibration.
    }
	\vspace{-10pt}
	\label{fig:frame_trajmatch}
\end{figure*}

The spatio-temporal calibration between roadside LiDARs intends to obtain the transformation parameters between LiDARs and align their timelines. Given the detection/tracking results from the point clouds, our task is to find the suitable transformation that facilitates the fusion of the data from different LiDARs coherently. Our system supports two data fusion modes. First, once calibrated, the raw point clouds can be directly merged to form a unified point cloud. Second, the transformation calculated by our algorithm can also support fusing the high level information, such as the features extracted by deep learning models, or even the object list and their trajectories. Fusion of raw point clouds, can maximally retain the information, which may potentially lead to fine-grained enhancements such as point cloud segmentation. However, this fusion method requires expensive hardware support, especially in terms of  network communication. Taking the RS-Ruby\_Lite~\cite{RS-LiDARLIte} as an example, output around $1,440,000$ points per second, around $30M$ of point cloud data are generated per second. As a result, in this work, we take the second approach and fuse the object-level information, which is a practical and effective approach given today's infrastructure capability.

Towards this goal, we propose \systemname  whose framework is shown in Figure~\ref{fig:frame_trajmatch}. Our calibration method does not assume any initial state of the system, neither the transformation parameters nor the time deviation between LiDARs. As a stateless system, we align the LiDARs both spatially and temporally at the same time, by going through an efficient iterative optimization process -- matching the object's positions, calculating the transformation parameters accordingly, and then aligning the trajectories with the parameters. We strive to make progress in each step, and in an iterative fashion converge to the optimal state. 
Below we briefly overview each step:
\vspace{-3pt}
\begin{enumerate}
    \item \vspace{4pt}\emph{Motion Feature Extraction.} After we obtain the trajectory information from each LiDAR that consists of a list of positions, we first extract the motion feature at each position, including the velocity and curvature. These two features are chosen for their rotation- and translation-invariant property.
    
    \item \vspace{4pt}\emph{Semantic Feature Matching}. Based on the similarity of motion features, we can obtain an initial set of position matches. However, many false matches appear due to the poor distinguishing power of motion features in our setting.   To filter out false matches, we adopt an array of more discriminative semantic information that has been generated in the object detection/tracking phase.

    \item \vspace{4pt}\emph{Iterative Optimization}. Next, we employ an iterative approach (similar to ICP [4]) to optimize the position matching and transformation parameter calculation at the same time.
    
    \item \vspace{4pt}\emph{Continuous Calibration}. Lastly, taking advantage of the relatively fixed location of roadside LiDARs, we propose a continuous calibration method. The core idea of this method is to continuously reduce errors by combining the calibration results of multiple passes.
    
\end{enumerate}

\section{SPATIO-TEMPORAL SYNCHRONIZATION METHOD}  

As shown in Figure~\ref{fig:frame_trajmatch}, \systemname consists of the three main steps: (i) point cloud based trajectory generation, (ii) trajectory motion feature extraction, (iii) semantics aware position matching, (iv) iterative optimization based calibration, and a method to further improve the success rate through continues calibrations. 
Below we discuss these steps one by one, using the calibration of two LiDARs as the running example.

\vspace{-0.3cm}
\subsection{Point Cloud based Trajectory Generation}

Different from the calibration algorithms that use raw point clouds as input, we use the point cloud detection/tracking results over a period of time $T$. We note that the choice of $T$ should be aware of the time offset between the two LiDARs. 

\begin{figure}[!t]
\centering
\begin{tabular}{cc}
\includegraphics[width=.45\linewidth]{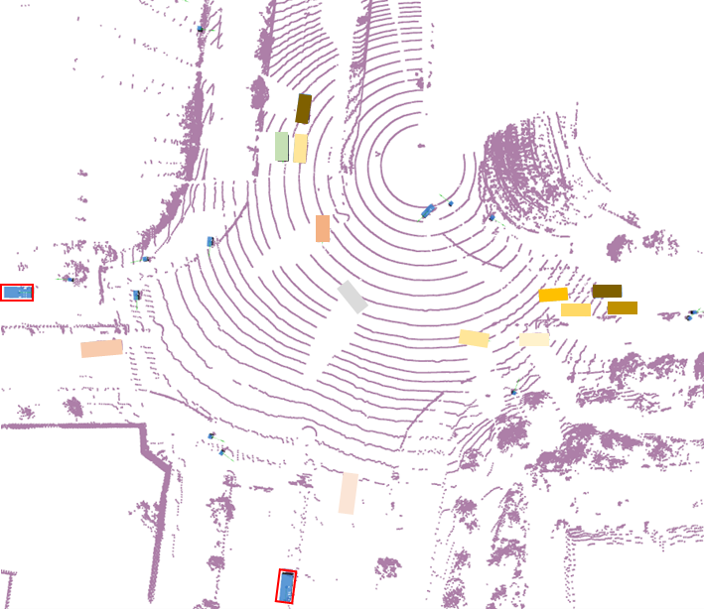}
& 
\includegraphics[width=.45\linewidth]{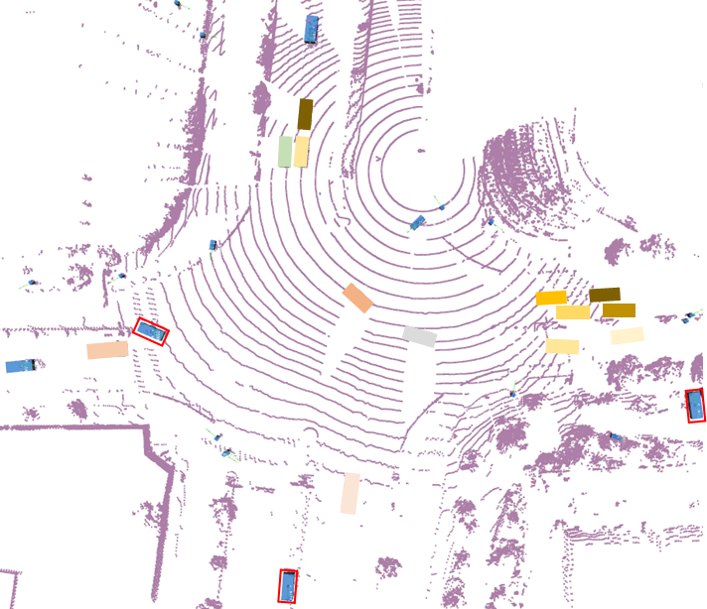} \\
(a) & (b)\\
\end{tabular}
\caption{ \textbf{An example detection and tracking result from two adjacent point cloud frames.} The detection algorithm is responsible for outputting the bounding box of the object, and the tracking algorithm is responsible for correlating the objects across frames. We use the same color to represent the same object, and we annotate mistakenly detected objects with red boxes. }
\label{fig:detection_tracking}

\end{figure}

As soon as the LiDAR generates a point cloud frame, the 3D object detection is performed. Many point cloud based  detection algorithms have been proposed recently, such as point-based algorithms in~\cite{pointnet} and voxel-based algorithms in~\cite{ deng2021voxel}.  These detection algorithms identify each object, classify it into one of the traffic participant categories such as cars, trucks, pedestrians, bicycles, etc., and estimate its 3D bounding box. Figure~\ref{fig:detection_tracking} gives an example point cloud frame and the objects detected from the frame using the PointPillars algorithm~\cite{pointpillars}. 

After performing object detection on each frame, we then perform the multi-object tracking (MOT) task to associate each object's positions across frames to obtain the movement trajectory. In this work, we engage the AB3DMOT algorithm~\cite{AB3DMOT}\cite{9839508} to calculate the trajectories. Figure~\ref{fig:detection_tracking} shows an example association result from two adjacent point cloud frames, in which we use the same color to denote the same car in these two frames. 

Here, the detection and tracking tasks constitute the preprocessing step for our \systemname system.

\vspace{-0.2cm}
\subsection{Trajectory Motion Feature Extraction}

Firstly, We follow the discussion in~\cite{survey_trajector_cluster} to give the mathematical definition of the trajectory. \\ 
A trajectory ($TR$) is a chronologically ordered sequence of 3D position coordinates of an object within a certain period of time, representing the motion information of an object, which is denoted by $ TR = \left\{ P_0, ..., P_i, ..., P_n \right\} (1 \leq i  \leq n)$. Here, $P_i$ is the position of the object at time $t_i$, where each position consists of a spatial coordinate set and a time stamp such as $ P_{i} = (x_i, y_i, z_i, t_i)$. To be more precise, position $P_i$  corresponds to the object's center coordinate ($(x_i, y_i, z_i)$) detected from point cloud  frame $f_i$ that is captured at time $t_i$. Furthermore, we call a set of trajectories a LiDAR detects a trajectory database ($TD$), with $TD = \left\{ TR_1,TR_2,TR_3,...,TR_m \right\}$, where $m$ is the number of objects that are detected. 

\begin{figure}[t]
\vspace{-0.3cm}
\centering
\begin{tabular}{cc}
\includegraphics[width=.75\linewidth]{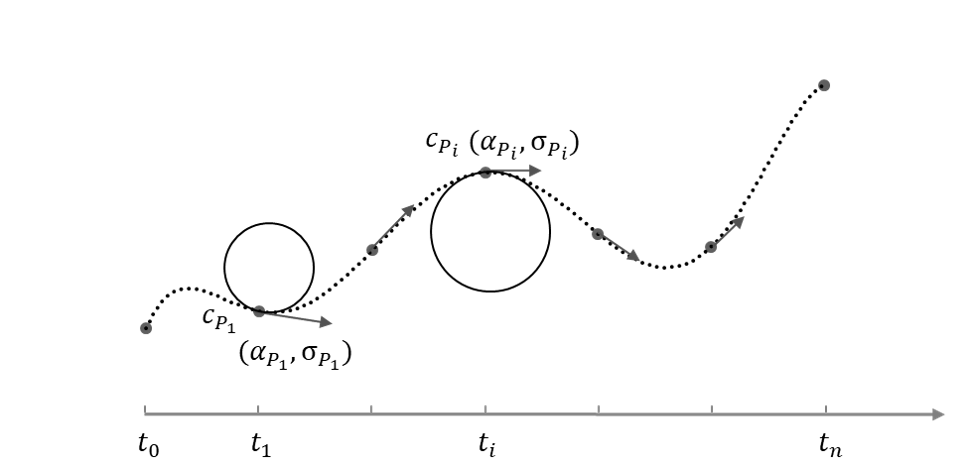}      
\end{tabular}
\caption{\textbf{Illustration of the trajectory motion features.}}
\label{fig:trajectory_feature}
\vspace{-0.2cm}
\end{figure}

Next, we extract suitable features for trajectories, which should accurately describe the object's motion pattern on the trajectory. Considering that a trajectory consists of a list of positions, we choose to extract motion features for each position to capture more fine-grained motion pattern. When choosing the features, we need to take into considerations the following factors. First of all, there exists an unknown coordinate transformation, the features need to be rotation-invariant and translation-invariant. Which means the feature extracted by algorithm should be consistent, regardless of how the coordinate is rotated and translated.  

Therefore, we use the velocity and curvature information at each position, both rotation-invariant, to describe the motion at that position. Suppose a trajectory is composed of $n$ positions. Then a trajectory's feature vector thus consists of $n$ velocity features and $n$ curvature features (as illustrated in Figure~\ref{fig:trajectory_feature}).

\begin{figure}[!t]
\centering
\begin{tabular}{cc}
\includegraphics[width=.45\linewidth]{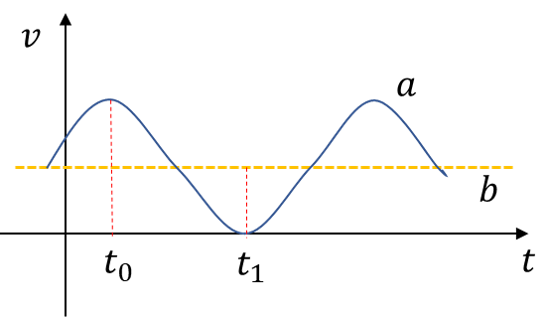}
& 
\includegraphics[width=.30\linewidth]{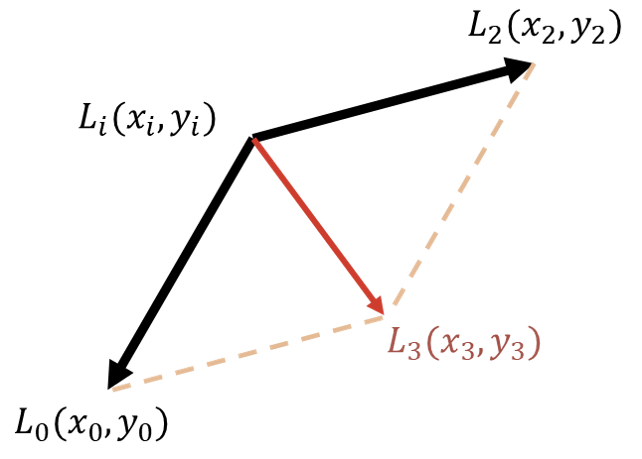} \\
(a) & (b)\\
\end{tabular}
\caption{\label{fig:cos_v}(a) trajectories with the same mean speed, but very different motion patterns, and (b) curvature calculation.}
\end{figure}

\vspace{4pt}\noindent\textbf{Velocity:} The velocity is an important feature for discriminating the mobility pattern.
First, we note that the mean speed alone (as in~\cite{TrajSimilarity}) is insufficient for this purpose. As shown in Figure~\ref{fig:cos_v}(a), during the interval from $t_0$ to $t_1$, even though trajectories $a$ and $b$ have the same mean speed, their movement patterns are quite different. As a result, we engage both the mean and  variance of the velocity here. 


For a trajectory $TR = \left\{ P_0, P_1, P2,...,P_n \right\}$,  we can get a sequence of velocities $\left\{ v_0, v_1,..., v_{n-1} \right\}$ for the trajectory positions, with $v_i= dis(P_i, P_i+1) / (t_{i+1} - t_i)$. Then the velocity feature $V(P_i)=\left\{ \alpha_i, \sigma_i \right\}$ for position $P_i$ can be calculated as:
\vspace{-4pt}
\begin{align*}
    \alpha(P_i) &= \bar{v} =  \frac{1}{2m} \sum_{j=i-m}^{j=i+m-1} v_j, \\ 
    \sigma^{2}(P_i) &= \frac{1}{2m} \sum_{j=i-m}^{j=i+m-1} {v_j - \bar{v}}^{2},
\end{align*}
where $m$ is a hyper-parameter to balance the detection/tracking noise. For exmaple, if the trajectory obtained from the tracker has significant noises, selecting a small $m$ likely leads to large errors in the feature estimation. 


\vspace{4pt}\noindent\textbf{Curvature:} The curvature characterizes the angle of a trajectory at a certain location. 
Suppose there are three consecutive positions: $P_{i-1}$, $ P_{i} $, $P_{i+1}$. The curvature at position $P_i$ is calculated as: 
\vspace{-4pt}
\begin{equation}
    c(P_i) = cos \theta = \frac{ \overrightarrow{P_{i} P_{i-1}} \cdot \overrightarrow{ P_{i} P_{i+1}} }{ \left| \overrightarrow{P_{i} P_{i-1}} \right| \left| \overrightarrow{P_{i} P_{i+1}} \right|},
    \ 
    \theta \in (0,\pi),
\end{equation}
where $ \theta $ denotes the angle between the two segments $\overrightarrow{P_{i-1} P_i}$ and $\overrightarrow{P_{i} P_{i+1}}$. We note that the smoother the trajectory is at position $P_i$, the closer $\theta$ is to $\pi$, and the smaller $c(P_i)$ is. 


\vspace{-0.3cm}
\subsection{ Semantics Aware Position Matching }

In this step, we establish the correspondences between two positions from different trajectories. 
First, we obtain the matching relationship of a series of positions based on the information of the position itself, and then we will consider the distribution of its neighbors to remove erroneous matches.

\vspace{4pt}\noindent\textbf{Motion-based Position Matching:} we utilize the Euclidean distance to measure the distance between two position motion features.  The distance between features $A=\left\{ c_A,\alpha_A, \sigma_A \right\}$ and $B\left\{ c_B,\alpha_B, \sigma_B \right\}$ is calculated as:
%
\begin{equation}
    d(A,B) = \lambda_c \left| c_A - c_B \right| + \lambda_{\alpha} \left| \alpha_A - \alpha_B \right| + \lambda_{\sigma} \left| \sigma_A - \sigma_B \right|,
    \label{eq:dis1}
\end{equation}
where the values of $ \lambda_c, \lambda_{\alpha}, \lambda_{\sigma} $  can be adjusted 
to give different weights to each of the features attributes. For example, between the velocity mean and variance, the variance $\sigma$ is more susceptible to noises. When the detection/tracking noise in the system is higher than a threshold, we can reduce $\lambda_{\sigma}$ to lower its weight.


Based on the above distance calculation, we next perform motion-based position matching. 
Suppose we have two LiDARs at an intersection, one with trajectory database $P$ and the other $Q$. Then for each position $p_i \in P$, we find the closest position $q_j$ in $Q$. If the $d(p_{i},q_{j}) < d_{th}$, we add the $(p_i, q_j)$ pair to the matched set. 

By going through the position matching algorithm, we obtain a large number of position pairs in which a position may have several matches. However, we find that many of these pairs are actually ``false'' matches. False matches can happen for the following reasons. Firstly, restricted by traffic rules, the speed and direction of movement of vehicles are mostly similar. Secondly, limited by the vehicle kinematics model, the speed and curvature of the vehicle usually change slowly~\cite{kong2015kinematic}, especially compared to the LiDAR sampling rate. Thirdly, there are errors in point cloud detection/tracking algorithms. All these reasons attribute to the high false match ratio. 

As an example, in a typical scenario with 2 LiDARs deployed at a 4-way intersection (see Figure~\ref{fig:low_overlap}), we obtain 875 matched position pairs, among which, only at most 175 pairs are actually valid.


\vspace{4pt}\noindent\textbf{Semantics Aware Position Matching:} \label{sec:rule}
We argue that the large number of false matches can be filtered out if we take into consideration the semantic information of the object, which has been generated in the earlier object detection and tracking tasks. 
In this work, we propose to explore the addition semantic information to enhance our position matching:  
\begin{itemize}
    \item \emph{Mutually Nearest Neighbor}. This is a simple observation. If positions $P_a \in TD_a$ and $P_b \in TD_b$ are a true match, then $P_a$ is the closest to $P_b$ in $TD_a$, and $P_b$ is the closest to $P_a$ in $TD_b$.
    
    \item \emph{Bounding Box Similarity}. If positions $P_a \in TD_a$ and $P_b \in TD_b$ are a true match, then their associated object bounding boxes (generated by the detector) should be similar to each other, including the box length, width, and height.   
    
    \item \emph{Neighbor Count Similarity}. If positions $P_a \in TD_a$ and $P_b \in TD_b$ are a true match, then the number of neighboring objects within a certain area is the same.

    \item \emph{Neighborhood Distribution Consistency}. If positions $P_a \in TD_a$ and $P_b \in TD_b$ are a true match, the distribution of the surrounding positions is similar and remains so in the adjacent frames. Based on this, we can construct a time-domain histogram to capture the distribution.
    
\end{itemize}

With these semantic information taken into consideration, we can effectively filter out the false matches. In the same example discussed above, the ratio of true matches increases from $22\%$ to $61\%$ after filtering. 


\vspace{-0.3cm}
\subsection{Iterative Optimization based Calibration}

Next, we considered the overall distribution of the data and employ an iterative approach (similar to ICP~\cite{ICP}) to optimize the position matches and to achieve accurate calibration at the same time. The optimization approach involves the three main operations: position matching, transformation parameter calculation, and trajectory matching. Namely, our optimization works as follows:

\begin{enumerate}
    \item \emph{$S1$: Transformation Parameter Calculation.} As discussed in Sec.~\ref{sec:background}, with a list of matched position pairs as the geometry constraints, we can calculate the transformation parameters by solving Eqn.~\ref{equ::least_sqrt}. Then the key to correct transformation parameters is to have a set of accurately matched position pairs.  
    
    \item \emph{$S2:$ Trajectory Matching.} Based upon the transformation parameters, we can convert each position's coordinates into a common system. With each position under the common coordinate system, we can now calculate the distance between two trajectories.
    In particular, we can obtain the matched trajectory pairs from matched position pairs and the trajectory information from the tracker. Meanwhile, we can also calculate the distance between matched trajectories. If this distance has become lower than the preset threshold, then we have achieved satisfactory calibration results and can terminate the optimization process.   
    
    \item \emph{$S3:$ Position Matching.} With the matched trajectory pairs and each trajectory's position sequence, we can further refine our matched position pairs by adding missed matches and removing false ones. Then we go back to $S1$, and continuously optimize until the distances between the matched trajectories are small than a threshold or the iteration exceeds the maximum number of times. 

\end{enumerate}

\vspace{-0.3cm}
\subsection{Continuous Calibration} \label{sec:continuous}

After discussing our calibration process, we next discuss how the system assesses the calibration effect and how we combine the effect of multiple calibrations. As such, we can achieve continuous calibration for the roadside LiDARs. 

Suppose our calibration results give the transformation parameters $\tau = ( R,T)$. 
%
%
%
%
Suppose $n_{pp}$ is the number of matched position pairs, $n_{po}$ is the number of positions in the overlapping area between the two LiDARs. Then we use the ratio $s=2*n_{pp}/n_{po}$ to roughly assess the score of the calibration operation.  A perfect calibration has score $s = 1$. 
In an actual system, we perform LiDAR calibration at a certain frequency. For example, we can perform calibration as soon as an object is detected, or every $k$ objects. In this process, we keep the current transformation parameters $\tau_{t}$ with a certain score $s_{t}$. In addition, suppose the latest calibration process gives $\tau_{t-1}$ and $s_{t-1}$. 

Then we adjust the transformation parameters as to achieve continuous calibration:
\begin{equation}
\tau = \cfrac{  {s_{t-1}}}{ {s_{t-1}} + s_t } \times {\tau_{t-1}}  +  \cfrac{s_{t}}{ {s_{t-1}} + s_t} \times \tau_t.
\label{equ::update}
\end{equation}

\begin{figure*}
\centering
\includegraphics[scale=0.6]{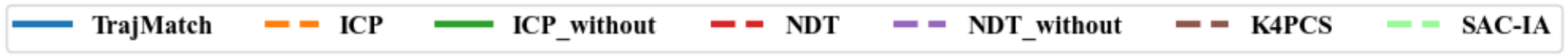}
\includegraphics[scale=0.5]{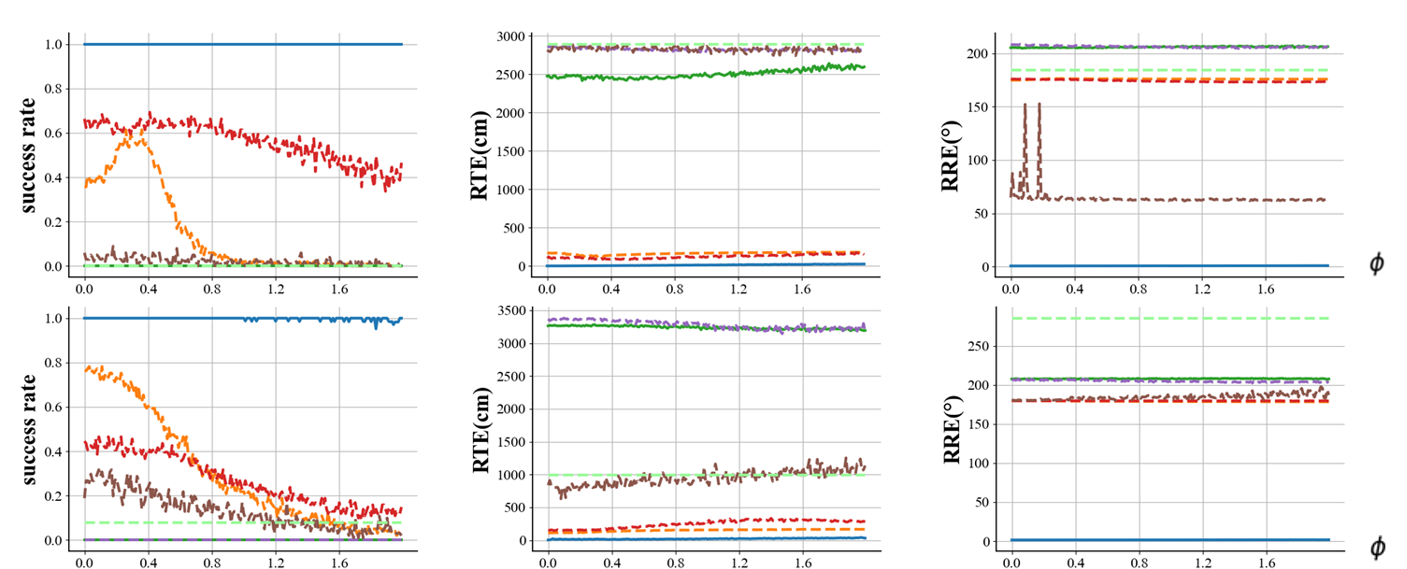}

\caption{\label{fig::result} We show here the effect of the algorithm in two different scenarios. The average value is used here, and our algorithm is filtered using our evaluation tool. In the first row, we put the LiDARs on the diagonal of the intersection, and in the second row, we put on the three-way intersection.}
\end{figure*}


\section{Performance Evaluation}\label{sec:results}   

In order to thoroughly evaluate \systemname,  we collect a simulated dataset \simdataset and a real-world dataset. The details of the dataset can be found in Section \ref{sec::dateset}. We define the evaluation metrics in Section \ref{sec::metrics}. In Section \ref{sec::result}, we show the evaluation results.

\vspace{-0.2cm}
\subsection{Evaluation Dataset}  \label{sec::dateset}
To validate the proposed method, we build two datasets, one real-world and one simulated.
 
\subsubsection{ \simdataset }

Since there is no public dataset to support the research on roadside LiDAR calibration, we construct \simdataset using the CARLA\cite{CARLA} and SUMO\cite{sumo} co-simulation platform.  There are several advantages of using a simulated dataset. Firstly, such a dataset can offer the ground truth information, including the translation of LiDARs, as well as the object detection/tracking information. Secondly, it is much easier to generate various traffic scenes for a much more thorough evaluation. 

Specifically, we simulate the traffic flow based on Sumo and then record the sensor data based on Carla. The following factors are considered when constructing this dataset. 
The first factor is the traffic flow, in which we add a different number of cars to the map to achieve different traffic flows. The second factor is the LiDAR type, including a 32-line LiDAR with a sensing range of 50m and an 80-line LiDAR with a sensing range of 150m. 


\subsubsection{Real-world Dataset} 

For the real-world dataset, we deploy two LiDARs in two scenarios and collect the data: (1) diagonally at the intersection with RS-Ruby Lite (80 beams); (2) on both sides of the sidewalk with RS-LiDAR-32 (32 beams).  
We use PointPillars\cite{pointpillars} as the detector and AB3DMOT\cite{AB3DMOT} as the tracker. 

\vspace{-2pt}
\subsection{Metrics and Baseline Algorithms}  \label{sec::metrics}

\subsubsection{Metrics:} 
To compare the performance with other calibration algorithms, we adopt the metrics defined in~\cite{RRE_RTE}, which are also used in~\cite{deep_global_regi, vi-eye}.  Firstly, the relative rotation error (\textbf{RRE}) is:
\begin{equation} 
    \begin{split}
    angle &= F \left(  {R_T^{-1} \cdot   R_E }  \right),   \\
    RRE &= \sum_{i=1}^3 \lvert   {angle(i)}    \rvert,
    \end{split}
\end{equation}
where $R_T$ and $R_E$ are the rotation matrices of the ground-truth transformation and the estimated transformation, respectively. $F( \cdot )$  transforms a rotation matrix to three Euler angles. Secondly, the relative translation error (\textbf{RTE}) is defined as:
\begin{equation}
    RTE = \left||  t_T - t_E  \right||,
\end{equation}
where $t_T$ and $t_E$ are the translation vector of the ground-truth transformation and the estimated transformation, respectively. 

In addition, we define a calibration session as a `success' when the RRE and RTE are below the predefined threshold~\cite{elbaz20173d}.  In this paper, we choose the threshold as $1m$, which is sufficient for the requirements of autonomous driving navigation~\cite{liu2018progress}. The \textbf{success rate} is then defined as the average rate a session is successful.

To test the algorithm's ability to time-synchronize, we used timing offset error (\textbf{TOE}). The TOE is defined as: 
\begin{equation}
    TOE = \left| T_T - T_E \right|,
\end{equation}
where $T_T$ and $T_E$ are the ground-truth and estimated timing offsets, respectively.

\subsubsection{Baseline Algorithms:} 
We compare \systemname with four baseline LiDAR calibration algorithms: (i) ICP~\cite{ICP}, in which the calibration results are achieved through iterative optimization,  (ii) NDT~\cite{ndt} in which the calibration is determined based on the similarity between the probability density functions (PDFs) of the point clouds; (iii) K-4PCS\cite{k4pcs} in which key points are extracted to improve the efficiency of the optimization process; (iv) SAC-IA\cite{SAC-IA}, in which the calibration is based on local geometry features around a point, called Fast Point Feature Histograms (FPFH). These algorithms are implemented based on the Point Cloud Library (PCL)~\cite{pcl}. We have carefully adjusted their parameter values to ensure they achieve the best performance with our datasets. Among these algorithms, only SAC-IA is based on raw point clouds, and the others based on detection/tracking results.

\begin{figure*}
\vspace{-0.4cm}
\centering
\includegraphics[scale=0.5]{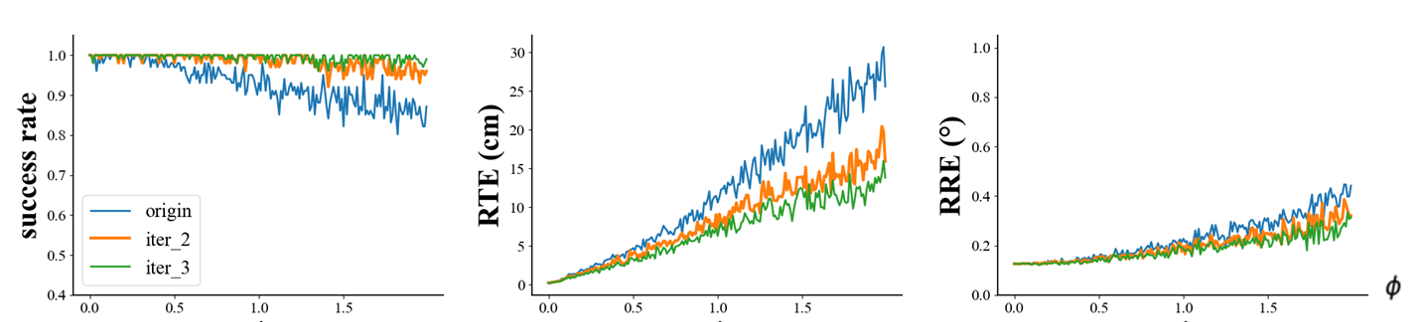}

\caption{\label{fig::result_four_iter} The spatial calibration results of after 1, 2, and 3 passes. For RTE and RRE.}
\vspace{-0.3cm}
\end{figure*}


\subsection{Calibration Results} \label{sec::result}
\subsubsection{Results with Simulated Dataset} We first show the results with the simulated dataset. 

\vspace{3pt}\noindent\textbf{Spatial Calibration vs Deployment Setting:}  In this set of experiments,  we compare \systemname with the baseline algorithms in terms of the spatial calibration results. To simulate different detection errors, we add the noise that follows the Gaussian distribution in the range $(0, \phi)$  to the ground-truth trajectory. We vary $\phi$ from 0 to 2m. According to the leaderboard of nuScenes\cite{nuscenes}, this value is significantly larger than the ranging error based on LiDARs detection, which is around 0.2m. By choosing a larger value, we make sure that our evaluation covers the worse case scenarios. In addition,
since the performance of ICP and NDT is largely dependent on the choice of the initial values, we test them under two situations, one with a hand-picked initial value and the other one without an initial value. 

The first row of figures~\ref{fig::result} show the success rate, RTE, and RRE at a 4-way intersection. The LiDARs are installed diagonally at the intersection with a distance of 28.8m, as shown in Figure~\ref{fig:install}(a). The number of LiDAR beam is 32, the detection distance is 50m. On these three metrics, our results have been consistently the best.  Even when $\phi$ is larger than 1.2m, \systemname can still achieve a success rate above 95\%. We also tested using 80-line 
LiDAR, and the results were similar to 32-line, so we didn't put result pictures here.  The second row show the result at a three-way intersection, and we came to a similar conclusion.

Furthermore, we demonstrate the effectiveness of the evaluation mechanism, through which we can easily catch these outliers and discount them. Table~\ref{tab:score} summarizes the scores for different cases and we are able to easily differentiate the outlier (with a score 0) and other normal cases (with score higher than 0.8). 

\begin{table}[tb]
    \vspace{-0.3cm}

    \small
    \centering
	\caption{The effectiveness of our parameter assessment.}

    \begin{tabular}{cccc}
	\toprule  & RTE(cm) & RRE($^{\circ}$) & score \\
	\midrule
         1 & 4.117 & 1.550 & 0.848   \\
         2 & 5.373 &  1.573 & 0.841  \\
         3 & 14.759 & 1.568 & 0.824  \\
         4 & 17.087 & 1.520 & 0.823  \\
         5 & 3569.470 & 183.878 & 0.00 \\
    
	\bottomrule
	\label{tab:score}
    \end{tabular}
    \vspace{-0.9cm}

\end{table}




\begin{table}[b]
\vspace{-0.3cm}

    \centering
    \small
	\caption{Spatial calibration results with different traffic flow. Here, we report the median values for RRE and RTE.}

	\begin{tabular}{ccccc}
	\toprule { Scenarios } & \makecell{ Number \\ of cars } & { RRE(${\circ}$) }  & { RTE(cm) } & \makecell{success \\ rate ($\%$)} \\ 
 
	\midrule
    \multirow{3}*{ \makecell[c]{four-way \\ intersection}} & 50 & 2.55 & 3.58 & 99 \\
        & 100 & 0.13 & 2.67 & 99 \\
        & 200 & 2.53 & 2.08 & 98 \\
	\bottomrule
	\label{table1}
    \end{tabular}
\end{table} 
\begin{table}[tb]
    \vspace{-0.5cm}

    \centering
    \small
	\caption{Spatial calibration results with different rotation angle. Here, we report the median values for RRE and RTE.}

	\begin{tabular}{ccccc}
	\toprule Scenarios & \makecell{Rotation \\ Angle($^{\circ}$)} & RRE($^{\circ}$) & RTE(cm) & \makecell{success \\ rate ($\%$)} \\
	\midrule
    \multirow{3}*{ \makecell[c]{four-way \\ intersection}} & 0 & 0.04 & 2.65 & 97 \\
        & 30 & 0.14 & 1.92 & 99 \\
        & 60 & 0.05 & 3.08 & 98.1 \\
        & 90 & 0.04 & 2.61 & 99 \\
        & 120 & 0.05 & 3.02 & 99 \\
    
	\bottomrule
	\label{table3}
    \end{tabular}
    \vspace{-0.5cm}
\end{table} 

\begin{figure}[!ht]
\vspace{-0.5cm}
\centering
\includegraphics[scale=0.4]{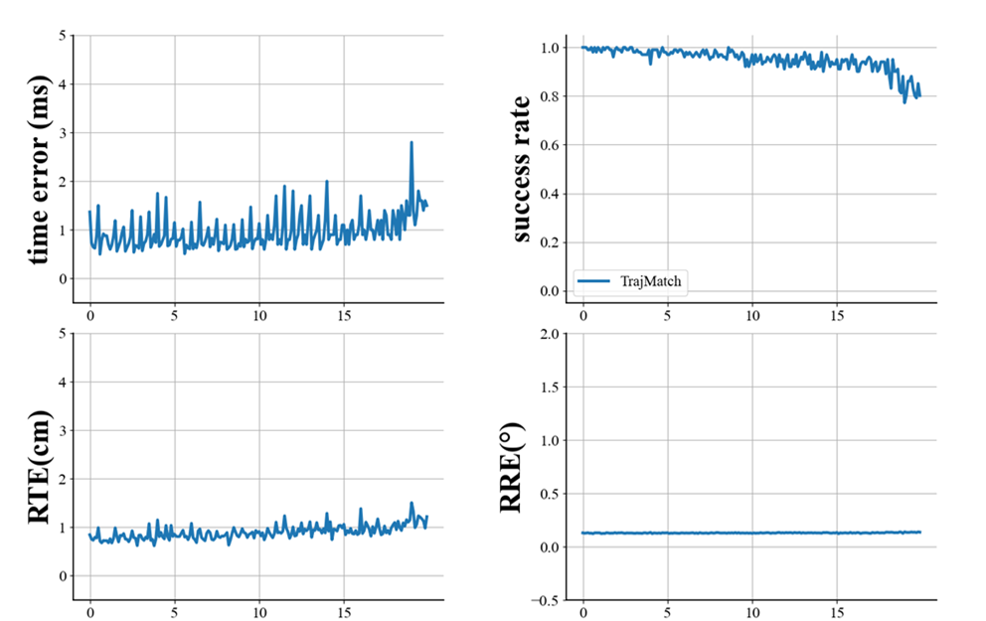}
\caption{\label{fig::o3_time_diff} Spatial-temporal calibration results with different time offset values at a 4-way intersection. The abscissa axis is the time deviation, from 0-20s. }
\vspace{-0.5cm}
\end{figure}

\def\svgwidth{\linewidth}
\begin{figure*}[!ht]
	\centering
 	\includegraphics[width=\linewidth]{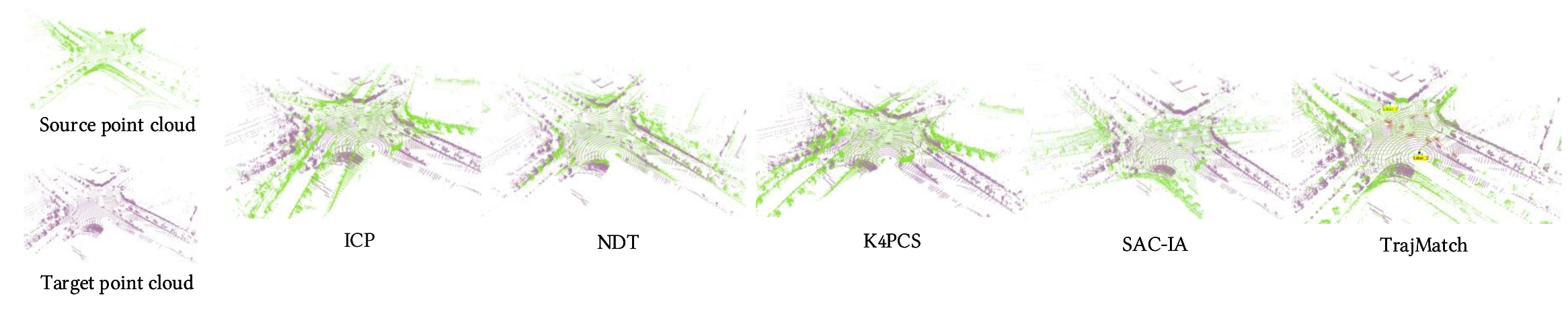}
	\caption{\textbf{Visualization results for different calibration algorithms with scenario 1. \systemname gives the best point cloud alignment.} }
	\vspace{-2pt}
	\label{fig:haikan_other_result}
	
	\centering
 	\includegraphics[width=1\linewidth]{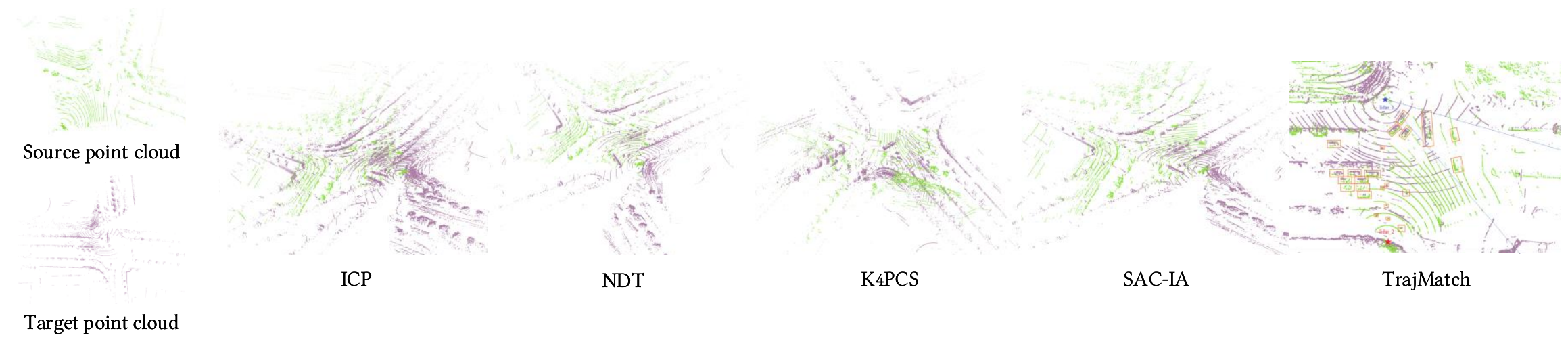}
	\caption{\textbf{Visualization results for different calibration algorithms with scenario 2. \systemname gives the best point cloud alignment.} }
	\label{fig::haikan_new_1}

\end{figure*}

\vspace{4pt}\noindent\textbf{Spatial Calibration vs Traffic Flow:}
In the second set of experiments, we test the effect of the traffic flow. Specifically, we simulate different traffic flows by randomly placing 50, 100, and 200 vehicles in the city. This parameter affects the number of cars that pass by our intersection and tests our capability of dealing with multiple cars at the same time. Here, we set $\phi = 0.2m$ and the time offset $ \Delta_t=0.5s$. The results are summarized in Table~\ref{table1}. We observe that the traffic flow does not have a noticeable impact on the performance. In all three situations, we have RTE less than $5cm$ and RRE less than $3$ degree. It suggests that \systemname can handle different traffic flows effectively.

\vspace{4pt}\noindent\textbf{Spatial Calibration vs Rotation:}
In the third set of experiments, we evaluate the robustness of the system to the LiDAR rotation angle. We vary the rotation angle while fixing the rest of the parameters: the time offset is $0.5s$, the standard deviation of the Gaussian noise  $\phi = 0.2m$. The calibration results are summarized in Table~\ref{table3}, which sufficiently demonstrate the our system is robust to different LiDAR rotation angles as we have chosen rotation-invariant features.  


\vspace{4pt}\noindent\textbf{Continuous Calibration:}
In the fourth set of experiments, we evaluate the effectiveness of our continuous calibration capability. That is, when we perform multiple rounds of calibration processes, can we improve the calibration accuracy?  We show the results in Figure~\ref{fig::result_four_iter} for the 4-way intersection case. 
The results show that continuous calibration can steadily improve the performance. In particular, when we have two rounds of calibration, the accuracy improves significantly compared to having a single round. However, the improvement becomes much less pronounced when we have three rounds or more. In fact, after three passes, the RTE decreases from 30cm to 10cm, and RRE decreases from 0.4° to 0.2°.  This renders \systemname a very viable approach that is able to refine its performance as more calibration sessions are conducted.

\vspace{4pt}\noindent\textbf{Temporal Calibration:}
Finally, we report the temporal calibration result. In this set of experiments, we have $\phi = 0.2m$. We vary the time deviations of the two LiDARs from 0 to 20s. Figures~\ref{fig::o3_time_diff} show the results for a 4-way intersection. As shown in Figure~\ref{fig::o3_time_diff}(d), When the time offset is less than 18s, we can achieve inter-frame synchronization, and the time synchronisation error is less than 2ms.  We believe that this is mainly due to the relative positions of vehicles that can form spatial constraints to assist our time calibration when there are multiple trajectories. Again, we believe such a temporal calibration accuracy is rather comparative for smart traffic applications. 


\subsubsection{Results with Real-World Dataset}
In addition to the simulated environment, we have also recorded the dataset in the real world for testing. We mainly tested two scenarios: (1) two LiDARs deployed diagonally at a large intersection, as shown in Figure~\ref{fig:install}(a), and (2) two LiDARs on both sides of a sidewalk, as shown in Figure~\ref{fig:install}(c).

Figure~\ref{fig:haikan_other_result} shows the combined point clouds for \systemname and 4 other algorithms. On this dataset, all 4 baseline algorithms fail to calibrate the LiDARs and align the point clouds. \systemname aligns the two point clouds nicely
. In this scenario,  the point clouds of the same object from different LiDARs are well matched. In addition, we also show the quantitative calibration results in Table~\ref{table2}. We observe that the translation error of the algorithm is less than $15cm$.


\begin{table*}[ht]
\vspace{-0.5cm}
    \small
    \centering
	\caption{The calibration results for real-world traces in scenario 1.}

	\begin{tabular}{c ccc cccccc}
	\toprule & \multicolumn{3}{c}{Angle error($^\circ$) } &  \multicolumn{6}{c}{Transform error} \\
	         & \multirow{2}{*}{$R_x$} & \multirow{2}{*}{$R_y$} & \multirow{2}{*}{$R_z$} & \multicolumn{2}{c}{$T_x$} & \multicolumn{2}{c}{$T_y$} & \multicolumn{2}{c}{$T_z$} \\ 
	        & & & & num(cm) & rate($\%$) & num(cm) & rate($\%$)  & num(cm) & rate($\%$) \\
	\midrule
        1 & -0.499 & 1.787 & -0.148 & 11.7 & 0.2 & 13.5 & 3.9 & 0.0 & 0.0 \\
        2 & -0.5 & 1.785 & -0.187 & 9.9 & 0.2 & 7.1 & 2.0 & 0.0 & 0.0 \\ 
         3 & -0.499 & 1.787 & -0.238 & 1.2 & 0.0 & 19.1 & 5.5 & 0.0 & 0.0 \\
        4 & -0.5 & 1.78  & 0.195 & 19.3 & 0.3 & 9.5 & 2.7 & 0.0 & 0.0 \\ 
	\bottomrule
	\label{table2}
    \end{tabular}
\vspace{-0.9cm}
\end{table*}

\def\svgwidth{\linewidth}




	


\section{RELEATED WORK} 

The spatio-temporal calibration of roadside LiDARs faces several important technical challenges due to the properties of the system deployment. Below we explain the challenges in details. 




\vspace{-0.3cm}
\subsection{Spatial Calibration:} The first challenge we face is the relatively small overlap ratio between two point clouds.
Specifically, the overlap ratio refers to the ratio between the intersection size and the union size. Let us look at a real-world deployment case (shown in Figure~\ref{fig:low_overlap}) with the following parameters: (1) the width of the 4-way intersection is $60m$, and (2) the detection range of LiDAR sensor mounted on vehicle is 80-100m (according to nuScenes~\cite{nuscenes}, with safety concerns in mind).  In this case, the overlap ratio is around $26\%$. 
Usually, LiDARs deployed on the roadside are tens of meters apart from each other to maximize the overall coverage, which might lead to an even smaller ratio in practice.

However, existing LiDAR calibration methods often assume a larger overlap ratio to achieve accurate results.  
For example, as discussed in~\cite{perfectMatch}, the state-of-the-art calibration methods require the two point clouds to overlap by at least 30$\%$ under the ground-truth transformation. Some algorithms require even larger overlap ratio; for example, ICP prefers the overlap ratio to be 80$\%$ or above~\cite{automated}. 

In addition, traffic monitoring systems usually have LiDARs installed at various heights (see Figure~\ref{fig:low_overlap}) . Due to the placement of the LiDARs, the resulting 3D overlap between two point clouds is almost reduced to a 2D area along the road surface, with a very short height~\cite{re_with_ground}.
Figure~\ref{fig:low_overlap} illustrates an overlap between two point clouds.   
Many existing calibration methods rely on discriminative hand-crafted features in the environment such as curves\cite{curves_feature}, intersection lines\cite{intersectionFeature}, adaptive covariance\cite{ZAI201715}, and it is hard to find these features in our scenarios. 

\vspace{-0.3cm}
\subsection{Temporal calibration:} 
For temporal calibration, we need to determine a constant time offset between sensors and align their timestamps~\cite{rehder2016general}. There are mainly two types of temporal calibration methods: ego-motion based and trajectory based. 

The first type of method, such as those discussed in~\cite{taylor2016motion, li2014online}, attempts to synchronize the timestamps based on the movement trajectories of the sensors. Specifically, this type of algorithm generally consists of two steps, estimating the ego-motion of each sensor and then calculating the time offset by aligning the trajectory based such as velocity\cite{qin2018online}, orientation change\cite{kelly2014general}. 
Obviously, this approach does not apply to our system because we have the LiDARs installed at fixed locations.

The second type of methods, such as those discussed in~\cite{pervsic2021spatiotemporal}, perform temporal and spatial calibration while having the sensors track the same target. It works in those settings where the target is known to all the sensors. However, in our system setting, LiDARs are usually deployed on busy roads where the sensors can detect many targets whose correspondences are unknown.  In addition, there are also methods that utilize additional clues for calibration, such as LiDARTag~\cite{lidartag}. They usually involve human involvement and are unsuitable for us.

\section{CONCLUSION}

In this paper, we present \systemname, a system that automates the spatio-temporal calibration for roadside LiDARs. Our calibration method does not assume any initial state of the system. As a stateless system, we align the LiDARs spatially with centimeter-level accuracy and temporally with millisecond accuracy, at the same time. The calibration is achieved through an efficient iterative optimization process. In addition, we also propose a continuous calibration mechanism that can accurately assess the quality of each calibration session and update the overall calibration when the assessment is above a certain threshold. 

We believe that traffic monitoring with roadside sensors will become an important component for future cities, and that our work takes the first step towards having a robust and accurate roadside perception system. Moving forward, we will continue to make effort to realize the vision. Specifically, we will investigate the calibration process for different deployment settings and different sensors. We will also carefully exploit the fused data from these sensors for improved traffic through and driving safety.  
\vspace{-0.3cm}


\bibliographystyle{IEEEtran}
\bibliography{reference}

\begin{thebibliography}{10}
\providecommand{\url}[1]{#1}
\csname url@samestyle\endcsname
\providecommand{\newblock}{\relax}
\providecommand{\bibinfo}[2]{#2}
\providecommand{\BIBentrySTDinterwordspacing}{\spaceskip=0pt\relax}
\providecommand{\BIBentryALTinterwordstretchfactor}{4}
\providecommand{\BIBentryALTinterwordspacing}{\spaceskip=\fontdimen2\font plus
\BIBentryALTinterwordstretchfactor\fontdimen3\font minus
  \fontdimen4\font\relax}
\providecommand{\BIBforeignlanguage}[2]{{%
\expandafter\ifx\csname l@#1\endcsname\relax
\typeout{** WARNING: IEEEtran.bst: No hyphenation pattern has been}%
\typeout{** loaded for the language `#1'. Using the pattern for}%
\typeout{** the default language instead.}%
\else
\language=\csname l@#1\endcsname
\fi
#2}}
\providecommand{\BIBdecl}{\relax}
\BIBdecl

\bibitem{arnold2020cooperative}
E.~Arnold, M.~Dianati, R.~de~Temple, and S.~Fallah, ``Cooperative perception
  for 3d object detection in driving scenarios using infrastructure sensors,''
  \emph{IEEE Transactions on Intelligent Transportation Systems}, 2020.

\bibitem{kim2021automated}
J.~Kim, C.~Kim, Y.~Han, and H.~J. Kim, ``Automated extrinsic calibration for 3d
  lidars with range offset correction using an arbitrary planar board,'' in
  \emph{2021 IEEE International Conference on Robotics and Automation
  (ICRA)}.\hskip 1em plus 0.5em minus 0.4em\relax IEEE, 2021, pp. 5082--5088.

\bibitem{xie2018infrastructure}
Y.~Xie, R.~Shao, P.~Guli, B.~Li, and L.~Wang, ``Infrastructure based
  calibration of a multi-camera and multi-lidar system using apriltags,'' in
  \emph{2018 IEEE Intelligent Vehicles Symposium (IV)}, pp. 605--610.

\bibitem{quenzel2016robust}
J.~Quenzel, N.~Papenberg, and S.~Behnke, ``Robust extrinsic calibration of
  multiple stationary laser range finders,'' in \emph{2016 IEEE International
  Conference on Automation Science and Engineering (CASE)}.\hskip 1em plus
  0.5em minus 0.4em\relax IEEE, 2016, pp. 1332--1339.

\bibitem{glas2015snapcat}
D.~F. Glas, D.~Br{\v{s}}{\v{c}}i{\v{c}}, T.~Miyashita, and N.~Hagita,
  ``Snapcat-3d: Calibrating networks of 3d range sensors for pedestrian
  tracking,'' in \emph{2015 IEEE international conference on robotics and
  automation (ICRA)}.\hskip 1em plus 0.5em minus 0.4em\relax IEEE, 2015, pp.
  712--719.

\bibitem{6385620}
K.~Schenk, A.~Kolarow, M.~Eisenbach, K.~Debes, and H.-M. Gross, ``Automatic
  calibration of a stationary network of laser range finders by matching
  movement trajectories,'' in \emph{2012 IEEE/RSJ International Conference on
  Intelligent Robots and Systems}, 2012, pp. 431--437.

\bibitem{pusztai2017accurate}
Z.~Pusztai and L.~Hajder, ``Accurate calibration of lidar-camera systems using
  ordinary boxes,'' in \emph{Proceedings of the IEEE International Conference
  on Computer Vision Workshops}, 2017, pp. 394--402.

\bibitem{xue2019automatic}
B.~Xue, J.~Jiao, Y.~Zhu, L.~Zhen, D.~Han, M.~Liu, and R.~Fan, ``Automatic
  calibration of dual-lidars using two poles stickered with retro-reflective
  tape,'' in \emph{2019 IEEE International Conference on Imaging Systems and
  Techniques (IST)}.\hskip 1em plus 0.5em minus 0.4em\relax IEEE, 2019, pp.
  1--6.

\bibitem{vi-eye}
Y.~He, L.~Ma, Z.~Jiang, Y.~Tang, and G.~Xing, ``Vi-eye: semantic-based 3d point
  cloud registration for infrastructure-assisted autonomous driving,'' in
  \emph{Proceedings of the 27th Annual International Conference on Mobile
  Computing and Networking}, 2021, pp. 573--586.

\bibitem{Radar-Camera}
Y.~Du, B.~Qin, C.~Zhao, Y.~Zhu, J.~Cao, and Y.~Ji, ``A novel spatio-temporal
  synchronization method of roadside asynchronous mmw radar-camera for sensor
  fusion,'' \emph{IEEE Transactions on Intelligent Transportation Systems}, pp.
  1--12, 2021.

\bibitem{WU2091105}
M.~Wu and B.~Coifman, ``Quantifying what goes unseen in instrumented and
  autonomous vehicle perception sensor data – a case study,''
  \emph{Transportation Research Part C: Emerging Technologies}, vol. 107, pp.
  105--119, 2019.

\bibitem{RS-LiDARLIte}
``Rs-lidar\_lite,'' \url{https://www.robosense.ai/rslidar/RS-Ruby_Lite},
  [n.d.].

\bibitem{pointnet}
C.~R. Qi and Su, ``Pointnet: Deep learning on point sets for 3d classification
  and segmentation,'' in \emph{CVPR}, 2017, pp. 652--660.

\bibitem{deng2021voxel}
J.~Deng, ``Voxel r-cnn: Towards high performance voxel-based 3d object
  detection,'' in \emph{Proceedings of the AAAI Conference on Artificial
  Intelligence}, vol.~35, no.~2, 2021, pp. 1201--1209.

\bibitem{pointpillars}
A.~H. Lang, S.~Vora, H.~Caesar, L.~Zhou, J.~Yang, and O.~Beijbom,
  ``Pointpillars: Fast encoders for object detection from point clouds,'' in
  \emph{CVPR}, 2019, pp. 12\,697--12\,705.

\bibitem{AB3DMOT}
X.~Weng, J.~Wang, D.~Held, and K.~Kitani, ``Ab3dmot: A baseline for 3d
  multi-object tracking and new evaluation metrics,'' 08 2020.

\bibitem{9839508}
Y.~Li, J.~Deng, Y.~Zhang, J.~Ji, H.~Li, and Y.~Zhang, ``${\mathsf{ezfusion}}$:
  A close look at the integration of lidar, millimeter-wave radar, and camera
  for accurate 3d object detection and tracking,'' \emph{IEEE Robotics and
  Automation Letters}, vol.~7, no.~4, pp. 11\,182--11\,189, 2022.

\bibitem{survey_trajector_cluster}
G.~Yuan, P.~Sun, J.~Zhao, D.~Li, and C.~Wang, ``A review of moving object
  trajectory clustering algorithms,'' \emph{Artificial Intelligence Review},
  vol.~47, no.~1, pp. 123--144, Jan 2017.

\bibitem{TrajSimilarity}
R.~S.~D. Sousa, A.~Boukerche, and A.~A.~F. Loureiro, ``Vehicle trajectory
  similarity: Models, methods, and applications,'' \emph{ACM Comput. Surv.},
  vol.~53, no.~5, 2020.

\bibitem{kong2015kinematic}
J.~Kong, M.~Pfeiffer, G.~Schildbach, and F.~Borrelli, ``Kinematic and dynamic
  vehicle models for autonomous driving control design,'' in \emph{2015 IEEE
  intelligent vehicles symposium (IV)}.\hskip 1em plus 0.5em minus 0.4em\relax
  IEEE, 2015, pp. 1094--1099.

\bibitem{ICP}
P.~Besl and N.~D. McKay, ``A method for registration of 3-d shapes,''
  \emph{IEEE Transactions on Pattern Analysis and Machine Intelligence},
  vol.~14, no.~2, pp. 239--256, 1992.

\bibitem{CARLA}
A.~Dosovitskiy, G.~Ros, F.~Codevilla, A.~Lopez, and V.~Koltun, ``Carla: An open
  urban driving simulator,'' in \emph{Conference on robot learning}.\hskip 1em
  plus 0.5em minus 0.4em\relax PMLR, 2017, pp. 1--16.

\bibitem{sumo}
D.~Krajzewicz, J.~Erdmann, M.~Behrisch, and L.~Bieker, ``Recent development and
  applications of sumo-simulation of urban mobility,'' \emph{International
  journal on advances in systems and measurements}, vol.~5, no. 3\&4, 2012.

\bibitem{RRE_RTE}
A.~Geiger, P.~Lenz, and R.~Urtasun, ``Are we ready for autonomous driving? the
  kitti vision benchmark suite,'' in \emph{2012 IEEE conference on computer
  vision and pattern recognition}.\hskip 1em plus 0.5em minus 0.4em\relax IEEE,
  2012, pp. 3354--3361.

\bibitem{deep_global_regi}
C.~Choy, W.~Dong, and V.~Koltun, ``Deep global registration,'' in
  \emph{Proceedings of the IEEE/CVF conference on computer vision and pattern
  recognition}, 2020, pp. 2514--2523.

\bibitem{elbaz20173d}
G.~Elbaz, T.~Avraham, and A.~Fischer, ``3d point cloud registration for
  localization using a deep neural network auto-encoder,'' in \emph{Proceedings
  of the IEEE conference on computer vision and pattern recognition}, 2017, pp.
  4631--4640.

\bibitem{liu2018progress}
J.~Liu, H.~Wu, C.~Guo, H.~Zhang, W.~Zuo, and C.~Yang, ``Progress and
  consideration of high precision road navigation map,'' \emph{Strategic Study
  of Chinese Academy of Engineering}, vol.~20, no.~2, pp. 99--105, 2018.

\bibitem{ndt}
P.~Biber and W.~Stra{\ss}er, ``The normal distributions transform: A new
  approach to laser scan matching,'' in \emph{Proceedings 2003 IEEE/RSJ
  International Conference on Intelligent Robots and Systems (IROS 2003)(Cat.
  No. 03CH37453)}, vol.~3.\hskip 1em plus 0.5em minus 0.4em\relax IEEE, 2003,
  pp. 2743--2748.

\bibitem{k4pcs}
P.~W. Theiler, J.~D. Wegner, and K.~Schindler, ``Keypoint-based 4-points
  congruent sets--automated marker-less registration of laser scans,''
  \emph{ISPRS journal of photogrammetry and remote sensing}, vol.~96, pp.
  149--163, 2014.

\bibitem{SAC-IA}
R.~B. Rusu, N.~Blodow, and M.~Beetz, ``Fast point feature histograms (fpfh) for
  3d registration,'' in \emph{2009 IEEE international conference on robotics
  and automation}.\hskip 1em plus 0.5em minus 0.4em\relax IEEE, 2009, pp.
  3212--3217.

\bibitem{pcl}
R.~B. Rusu and S.~Cousins, ``3d is here: Point cloud library (pcl),'' in
  \emph{2011 IEEE international conference on robotics and automation}.\hskip
  1em plus 0.5em minus 0.4em\relax IEEE, 2011, pp. 1--4.

\bibitem{nuscenes}
``Leaderboard of detection on nuscenes,''
  \url{https://www.nuscenes.org/object-detection?externalData=all&mapData=all&modalities=Any},
  [n.d.].

\bibitem{perfectMatch}
Z.~Gojcic, C.~Zhou, J.~D. Wegner, and A.~Wieser, ``The perfect match: 3d point
  cloud matching with smoothed densities,'' in \emph{Proceedings of the
  IEEE/CVF Conference on Computer Vision and Pattern Recognition}, 2019, pp.
  5545--5554.

\bibitem{automated}
P.~Kim, J.~Chen, and Y.~K. Cho, ``Automated point cloud registration using
  visual and planar features for construction environments,'' \emph{J. Comput.
  Civ. Eng}, vol.~32, no.~2, p. 04017076, 2018.

\bibitem{re_with_ground}
\BIBentryALTinterwordspacing
R.~Yue, H.~Xu, J.~Wu, R.~Sun, and C.~Yuan, ``Data registration with ground
  points for roadside lidar sensors,'' \emph{Remote Sensing}, vol.~11, no.~11,
  2019. [Online]. Available: \url{https://www.mdpi.com/2072-4292/11/11/1354}
\BIBentrySTDinterwordspacing

\bibitem{curves_feature}
B.~Yang and Y.~Zang, ``Automated registration of dense terrestrial
  laser-scanning point clouds using curves,'' \emph{ISPRS journal of
  photogrammetry and remote sensing}, vol.~95, pp. 109--121, 2014.

\bibitem{intersectionFeature}
I.~Stamos and M.~Leordeanu, ``Automated feature-based range registration of
  urban scenes of large scale,'' in \emph{2003 CVPR}, vol.~2.\hskip 1em plus
  0.5em minus 0.4em\relax IEEE, 2003, pp. II--Ii.

\bibitem{ZAI201715}
D.~Zai and J.~Li, ``Pairwise registration of tls point clouds using covariance
  descriptors and a non-cooperative game,'' \emph{ISPRS Journal of
  Photogrammetry and Remote Sensing}, vol. 134, pp. 15--29, 2017.

\bibitem{rehder2016general}
J.~Rehder, R.~Siegwart, and P.~Furgale, ``A general approach to spatiotemporal
  calibration in multisensor systems,'' \emph{IEEE Transactions on Robotics},
  vol.~32, no.~2, pp. 383--398, 2016.

\bibitem{taylor2016motion}
Z.~Taylor and J.~Nieto, ``Motion-based calibration of multimodal sensor
  extrinsics and timing offset estimation,'' \emph{IEEE Transactions on
  Robotics}, vol.~32, no.~5, pp. 1215--1229, 2016.

\bibitem{li2014online}
M.~Li and A.~I. Mourikis, ``Online temporal calibration for camera--imu
  systems: Theory and algorithms,'' \emph{The International Journal of Robotics
  Research}, vol.~33, no.~7, pp. 947--964, 2014.

\bibitem{qin2018online}
T.~Qin and S.~Shen, ``Online temporal calibration for monocular visual-inertial
  systems,'' in \emph{2018 IEEE/RSJ International Conference on Intelligent
  Robots and Systems (IROS)}.\hskip 1em plus 0.5em minus 0.4em\relax IEEE,
  2018, pp. 3662--3669.

\bibitem{kelly2014general}
J.~Kelly and G.~S. Sukhatme, ``A general framework for temporal calibration of
  multiple proprioceptive and exteroceptive sensors,'' in \emph{Experimental
  Robotics}.\hskip 1em plus 0.5em minus 0.4em\relax Springer, 2014, pp.
  195--209.

\bibitem{pervsic2021spatiotemporal}
J.~Per{\v{s}}i{\'c}, L.~Petrovi{\'c}, I.~Markovi{\'c}, and I.~Petrovi{\'c},
  ``Spatiotemporal multisensor calibration via gaussian processes moving target
  tracking,'' \emph{IEEE Transactions on Robotics}, vol.~37, no.~5, pp.
  1401--1415, 2021.

\bibitem{lidartag}
J.-K. Huang and Wang, ``Lidartag: A real-time fiducial tag system for point
  clouds,'' vol.~6, no.~3, pp. 4875--4882, 2021.

\end{thebibliography}

\end{document}